# Using CNNs For Users Segmentation In
# Video See-Through Augmented Virtuality

Pierre-Olivier Pigny
CLARTE
Laval, France
pigny@clarte-lab.fr

Lionel Dominjon
CLARTE
Laval, France
dominjon@clarte-lab.fr

*Abstract*— In this paper, we present preliminary results on the use of deep learning techniques to integrate the user's self-body and other participants into a head-mounted video see-through augmented virtuality scenario. It has been previously shown that seeing user's bodies in such simulations may improve the feeling of both self and social presence in the virtual environment, as well as user performance. We propose to use a convolutional neural network for real time semantic segmentation of users' bodies in the stereoscopic RGB video streams acquired from the perspective of the user. We describe design issues as well as implementation details of the system and demonstrate the feasibility of using such neural networks for merging users' bodies in an augmented virtuality simulation.

*Keywords – augmented virtuality, semantic segmentation, video see-through*

## I. INTRODUCTION

Augmented Virtuality (AV) is a sub-category of mixed reality environments where real world objects are integrated into a computer generated environment [1]. In particular, users' bodies can be merged into the simulation in order to increase their feeling of self [2], [3] and social presence [4]. Self-embodiment also facilitates the perception of spatial characteristics of the virtual world such as distance [5] and position [6] estimation. In particular, high fidelity visual hand feedback has been demonstrated to produce a strong visuo-proprioceptive integration [7] and shorter movement times compared to a lower fidelity hand feedback [8].

Several approaches have been proposed to date to integrate users' bodies in an AV simulation. Most of them rely on computer vision techniques to reconstruct the body either through point clouds [9], polygon meshes [10], or voxels [3] models. Reconstructing real users' bodies to integrate them into the simulation has the main advantage of providing a view-independent representation. But except for a couple of high-end complex systems [10], they also usually suffer from the improving but still rather poor quality of the depth maps provided by current hardware [11]. In addition, most reconstruction-based systems rely on said outside-in setup [3], [9] (as opposed to head-mounted, inside-out, systems [12]) that may lead to occlusions preventing a whole reconstruction of the users. Other approaches provide depth perception without reconstruction steps by presenting synchronized stereoscopic video pairs to the user's eyes.

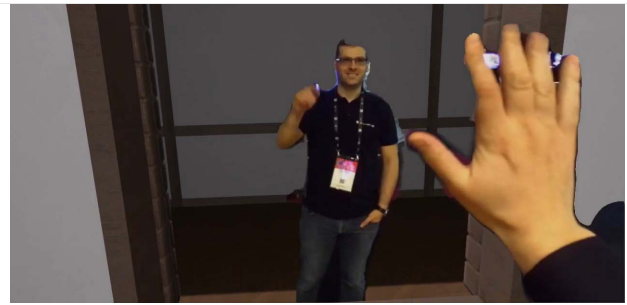

Fig. 1. Self-body and collaborator's body viewed from the user perspective and merged into an augmented virtuality experience.

That's the underlying concept of head-mounted Video See-Through (VST) approaches [13]: cameras mounted on the headset close to the user's eyes capture the real world from the perspective of the user, while the position and the orientation of the tracked head is used to render the Virtual Environment (VE). Using a VST approach for AV has the main advantage of not being impacted by depth map quality, as it is not required actually. In addition, because the reconstruction step is not necessary, VST approaches are also faster.

But video streams segmentation is required to isolate users' bodies from other objects. The particularity of the segmentation required for merging bodies into a VST-based AV simulation is that 1) bodies can look very different depending on the point of observation (user's self-body vs. others' bodies as observed from the perspective of the user) and nevertheless they should be identified as the same "person" class and 2) the segmentation process has to be fast enough for keeping up with update rates required for immersive experiences [14], or at least with the cameras update rate. To our knowledge, solutions that have been proposed to date rely exclusively on chroma-keying, which suffers from significant limitations. Recent advances in deep learning techniques, and in particular on Convolutional Neural Networks (CNNs)-based semantic segmentation, make us wonder whether they may be used in a VST-based AV scenario. In particular, can CNNs-based semantic segmentation be both accurate and fast enough for an AV experiences? Can it be versatile enough to segment bodies observed from both an ego- and exocentric point of view in an acceptable time scale? What training set should be used?

In this paper, we present preliminary work and demonstrate the feasibility of using CNNs for segmenting human bodies in VST video feeds captured from the user perspective (see sample result in Fig. 1) in real-time and report on implementation details used to achieve such a result. The paper is organized as follows. Section II describes related work in terms of body segmentation in live video streams. Section III provides an insight on semantic segmentation design and implementation and measured performance. Section IV discusses the results, with a focus on qualitative evaluation. Finally, we summarize our conclusions in section V along with mentioning our on-going and future work.

## II. RELATED WORK

To our knowledge, all the proposed solutions to date to integrate selected objects in a VST-based AV simulation rely on chroma keying. Chroma keying refers to a technique for compositing two video streams together based on color hues. Implementations differ in the sense of what is represented by the keying color: either the pixels to be discarded [15], or the pixels to be composited [2], [16]–[18]. The former is the most widespread approach and have been used for decades in the cinema industry. Its main drawback is that it requires dedicated facilities ("green screen studios"), which makes it incompatible with many applications, especially for home usage. The opposite technique may seem more convenient as it does not impose any constrain on the user physical environment, but it will fail at segmenting a whole body composed of many different colors (hair, clothes, shoes…).

Beyond the scope of VST display, other techniques for segmenting human bodies in a real-time video stream have been proposed. Some methods use randomized decision forests together with hand-crafted image features to segment either hands or full body in a video stream [19]–[22]. The latter implementation has been popularized through its integration in the official Microsoft Kinect SDK. Nevertheless, it seems to only support quasi-static backgrounds and would not work with a sensor bound to the head of the user. Finally, recent work also investigates deep learning approaches, either on a depth map [23] or on an RGB map [24].

## III. SEMANTIC SEGMENTATION FOR AV EXPERIENCES

In the following we describe the CNN model we use to perform semantic segmentation in the cameras streams. We also report implementation details and provide an evaluation of the performances of our technique.

### A. Design of the semantic segmentation model

The semantic segmentation model is designed with inference speed as a leading constraint to limit the overhead due to segmentation in the whole rendering process of the AV scenario and keep a comfortable experience.

#### 1) Network architecture

The selected architecture is based on a modified U-net [25] CNN architecture. U-net extends Fully Convolutional Networks (FCN) [26] by requiring much smaller training datasets and yield more precise segmentations. In addition,

U-net architecture is reputed to be fast enough to perform some basic segmentation tasks at 30Hz on mobile platforms [27], which makes it a relevant candidate for our application. It is also now a well-known, efficient, and widely used architecture. In the rest of the paper our network will be referred as SSTU-net, standing for Semantic See-Through U-net. SSTU-net is made of two parts, a 5-layer FCN encoder with pooling which extracts features from an RGB image, and a U-net-like 5-layer FCN decoder with up-sampling deconvolution and concatenation layers. Each layer contains two 3x3 convolutions, batch normalization and ReLU activation and one 2x2 pooling or up-sampling layer. The last 1x1 convolution decoder layer is replaced by a sigmoid activation function to obtain a segmentation probability instead of an absolute decision. This feature will be used during compositing to reduce flickering and soften contours.

#### 2) Datasets

For training our model, we used the COCO dataset [28] as it is one the most comprehensive datasets currently available for semantic segmentation. Since we're only interested in body segmentation, COCO was restricted to images containing instances of the "person" class, referenced as COCO-body in the rest of this paper. Because COCO-body does not include any human body seen from an egocentric point of view, we had to complete COCO-body with our own custom dataset we called EGO-body. EGO-body was created using a chroma-keying technique similar to [29] to capture bodies viewed from an egocentric point of view and the associated ground-truth segmentation mask.

Images were captured on a green background through a stereoscopic head-mounted camera. Seven people were asked to look around while wearing the camera and shoot pictures showing their body in as many different conditions as possible. The height and the orientation of the head while shooting were also collected. Every non-green pixel was considered as the ground-truth mask. In addition, a collection of backgrounds (i.e. pictures with no human body part) was shot from the same height and orientation, and then composited with the foreground. EGO-body contains about 1500 train images and 120 validation images.

#### 3) Training

Weights are initialized by training a base version of SSTU-net using COCO-body with a batch size of 12 and an Adam optimizer. The resulting model is referred as SSTU_coco. Learned weights are then fine-tuned with a balanced dataset containing all EGO-body data and a random subset of COCO-body. Two fine-tuned models were trained: one with fine-tuning performed on the decoder only (SSTU_f1), with fixed encoder weights, and one with fine-tuning performed on both the encoder and the decoder (SSTU_f2).

As EGO-body is a rather small dataset, a lot of data augmentation was performed during the training phase. Different random transformations were applied on the training set regarding brightness, saturation, hue, contrast, normal distributed noise, horizontal flip, image scaling (up to 10%) and image rotation. Image rotation was bound to [-30°; +30°] and vertical flip was not used to remain consistent with our head-worn camera use-case.

## B. Implementation of the model

### 1) Hardware setup

Our hardware setup consists of a PC (Intel Core i7-7800X CPU - RAM 64GB - two Nvidia GeForce GTX 1080Ti graphics cards) and a ZED mini stereoscopic camera mounted on an HTC VIVE headset (see Fig. 2). The ZED mini camera provides two synchronized 720p@60Hz video streams with an average latency of 37ms [30]. It is worth noting that VST headsets equipped with high performance embedded cameras will be available soon [31], thus probably removing the need for any additional hardware for our technique in a near future.

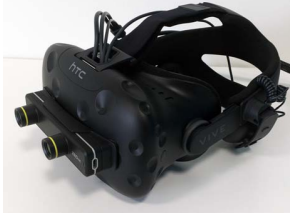

Fig. 2. ZED mini stereoscopic camera mounted on an HTC VIVE headset.

### 2) Software implementation

The segmentation module is implemented on top of Google TensorFlow and is fed with images captured from the ZED mini camera and downscaled to 256x256 pixels which seems to be an acceptable processing speed / accuracy tradeoff. Processing of both eyes is parallelized on the two graphics cards. The output of the module is a 256x256 segmentation mask for each eye, i.e. a probability map for each pixel to belong to the "person" class. The mask is then set as an alpha channel for the displayed video stream, resulting in displaying uncertain pixels as transparent, with a level of transparency depending on the probability. This helps reducing flickering a lot.

To render the video streams in the VR headset, the Zed mini camera is modelized as two virtual cameras cam$_{Zed}$, while the eyes of the user are modelized as two other virtual cameras cam$_{Eyes}$. Cam$_{Zed}$ and cam$_{Eyes}$ projection matrices are derived from the intrinsics of the Zed mini camera provided by the vendor and the field of view of the VR headset respectively. To address the difference of field of view, the rendering region is restricted to the overlap between cam$_{Zed}$ and cam$_{Eyes}$. In other words, the content outside of cam$_{Zed}$ frustum is not rendered. Both images of the stereoscopic video stream are mapped on planes located in front of cam$_{Zed}$ and covering its whole field of view. The rendering of the planes is performed through a shader that makes use of the depth map provided by the Zed mini camera to perform the z-test for every pixel of the plane. This allows to handle occlusions between the video stream and the 3D content (see Fig. 6. bottom-center).

To deal with the delay introduced along the image workflow – from capture to presentation - a Camera Time Warp [32] algorithm has been implemented. This ensures video frames are consistent and synchronized with the computer generated content. In our case, the system has to compensate for the camera capture latency plus the overhead for segmentation inference time and data transfer.

### 3) Performances

*Execution time*. With our implementation, the whole video processing runs in an average 22ms (data preparation 6ms + inference time 16ms), i.e. it can be run at a 45Hz frequency. Including the camera capture latency, frames can thus be delivered to the display system with an average delay of 59ms.

*Segmentation quality*. To measure the quality of the segmentation, we extracted a subset of 500 random images from COCO-body train part and 500 images from EGO-body. All the images in these test sets were not used for training the model. Since ground-truth data are represented by a binary mask, computing a mean Intersection over Union (mIoU) indicator requires to binarize predictions, which is achieved using a 0.5 threshold. The mIoUs computed for every model based on SSTU-net are summarized in Table I for the two test sets. The first line shows that the model trained on COCO-body only, SSTU_coco, does not perform well on EGO-body data, as expected. Unfortunately, the fine-tuned models SSTU_f2 and SSTU_f1 do not perform better since their performances, on COCO-body this time, drop off. We suspect that, because body shapes in the two datasets are very different, the features learned by the decoder might be in concurrence.

TABLE I. SEGMENTATION ACCURACY (mIOU)

|  | COCO-body | EGO-body |
|---|---|---|
| SSTU_coco | 0.63 | 0.52 |
| SSTU_f2 | 0.33 | 0.73 |
| SSTU_f1 | 0.42 | 0.50 |

## C. Optimization

Following this hypothesis, our model has been modified to test a two-decoder approach. In this new architecture, referred as SSTU-net-2, the encoding part remains the same, but two decoders are running in parallel: one is in charge of computing the probability for every pixel to belong to a "person-exo" class, and the other one predicts the belonging to a "person-ego" class (see Fig. 3). The probability for a pixel to belong to the integrated "person" class is then computed as the maximum value of both predictions. The encoder-decoder path used to predict "person-exo" pixels is referred as SSTU_exo, while the path used to predict "person-ego" pixels is referred as SSTU_ego. The model aggregating SSTU_exo and SSTU_ego is referred as SSTU. SSTU_exo is basically the same as SSTU_coco. Then, the decoder weights of SSTU_ego are computed by freezing SSTU_exo weights and training the model on the balanced dataset. In order to specialize SSTU_ego, the COCO-body data part of the balanced dataset is considered as background. With this new implementation, the whole video processing runs in an average 29ms (data preparation 6ms + inference time 23ms) and can thus be run at 34Hz. That means that currently only 1 frame out of 2 can be processed in the video stream (i.e. 1 out of 2 is dropped), which is not ideal but enough to demonstrate the feasibility and the interest of the technique. The delay to the display system thus raises to 66ms.

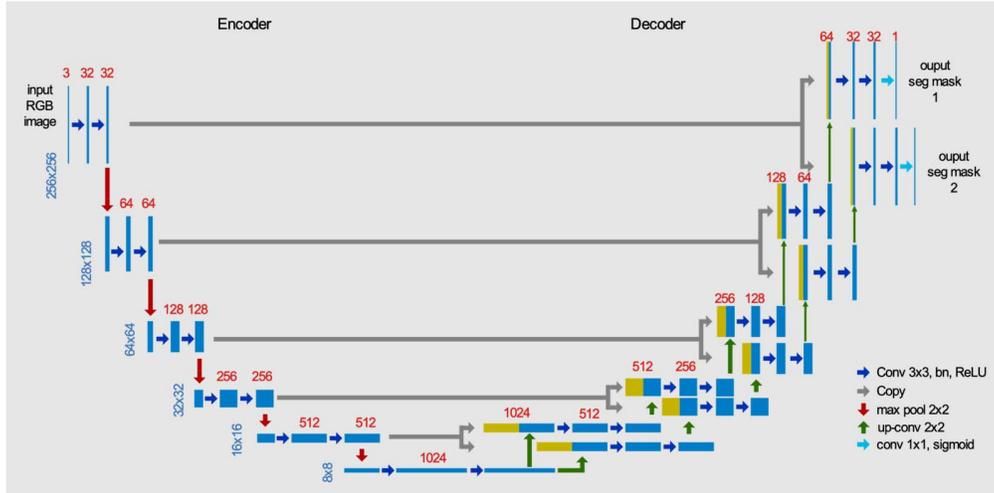

Fig. 3. SSTU-net-2 architecture: 1 encoder path, 2 decoder paths (person-exo, person-ego).

As stated above, SSTU_exo is identical to SSTU_coco, so their quality is the same as well: it performs well on exocentric points of view but is limited on egocentric points of view (see Table II). Unsurprisingly, the exact opposite is observed for SSTU_ego. But in the end, we can observe that the aggregated model, SSTU, succeeds in remaining as accurate as SSTU_exo on exocentric data while improving its accuracy on egocentric point of views.

TABLE II. TWO-DECODER SEGMENTATION ACCURACY (MIoU)

|  | COCO-body | EGO-body |
|---|---|---|
| SSTU_exo | 0.63 | 0.52 |
| SSTU_ego | 0.01 | 0.86 |
| SSTU | 0.63 | 0.69 |

## IV. DISCUSSION

Fig. 5 illustrates SSTU quality from a qualitative point of view. The image samples are extracted from the test sets. For each original RGB image (left column), ground truth (middle column) and SSTU prediction (right column) are presented. Three rows of images illustrate different levels of segmentation quality in terms of IoU for each of the two datasets: the first two rows illustrate what may considered as "good" quality segmentation (image #1: IoU = 0.94, image #2: IoU = 0.95), while the last two rows may be considered as "bad" (image #3: IoU = 0.17, image #4: IoU = 0.46).

Yet, informal preliminary tests tend to show that the mIoU metric cannot reflect on its own the quality of the experience in a VST AV scenario. For example, the IoU of image #3 is very low (0.17), but these segmentation errors would probably not be disturbing in an AV scenario: the person in the background is almost completely missed by SSTU, hence the low IoU, but is represented by only a small number of pixels. This error will probably not be noticed by the user and its impact on the user experience will be low. On the other hand, the IoU of image #4 is much higher (0.46), but the proportion of pixels incorrectly classified as person looks so high that the user experience will be greatly impaired. Therefore, in the case of an AV scenario, we believe that the Pixel Accuracy (PA) seems to be a metric that can better reflect the quality of the user experience. Let's get back to image #3 example: PA of the prediction is 0.99, which means that most of the pixels are correctly classified. The impact of incorrectly classified pixels on the user experience will be low. Regarding image #4, PA is equal to 0.7, which means that 30% of the pixels are incorrectly classified: this can have a huge impact on the user experience, considering that the images cover a large portion of the user field of view - despite a greater IoU. On our test sets, the global mean PA for COCO-body is $mPA_{COCO-body} = 0.96$, and equal to $mPA_{EGO-body} = 0.92$ for EGO-body, which means that, in average, a very large proportion of pixels is correctly classified by SSTU (see samples in Fig. 6, upper row). This seems consistent with the good feedbacks we received from our informal evaluations.

A more in-depth observation of the results seems to indicate that SSTU tends to classify as person more pixels than it is supposed to (see Fig. 6, bottom right sample). To confirm this observation, pixel precision and recall indicators for the person class have been computed. On the EGO-body test set, they show that almost all person pixels in the images have been correctly classified ($recall_{EGO-body} = 0.93$), but also that a significant amount of background pixels has also been classified as person ($precision_{EGO-body} = 0.74$). Regarding COCO-body test set, a lot of person pixels may have been missed ($recall_{COCO-body} = 0.68$), but pixels are more reliably classified as person than on EGO-body ($precision_{COCO-body} = 0.85$). Overall, the low values of precision on both datasets indicate that background pixels can sometimes be incorrectly classified as person pixels by SSTU. We hypothesize that this may be due to the lack of images with no person in our training sets. Possibly, SSTU has not correctly learned how not to detect persons.

Finally, our tests show that the segmentation consistency between stereo pairs can sometimes be visible enough to impair the user experience (see Fig. 4). This confirms that this issue should be addressed to support the adoption of our technique.

## V. CONCLUSION AND FUTURE WORK

We have presented preliminary work that demonstrates the feasibility of using CNNs for semantic segmentation of users' bodies in a head-mounted VST-based AV simulation. Our approach allows for real-time segmentation of bodies seen from both exo- and egocentric point of view at the same time. Even if in this preliminary implementation, segmentation execution speed can't currently totally keep up with the cameras update rate while maintaining a decent quality, results are promising and already sufficient to demonstrate the interest of the approach. In addition, the rapid evolution of CNNs research [33]–[35] suggest that this issue will probably be solved soon. We will not fail to test and compare these new network structures in the near future. Hybrid approaches [36] will also be another option to speed up the segmentation process. Another direction in our future work will deal with stereoscopic consistency, which we have not addressed yet. We'll indeed need to find a way to ensure that masks computed by semantic segmentation are consistent between both eyes. Finally, despite our focus on body segmentation in this paper, our technique could be extended to any other class of objects: keyboards, smartphones, furniture… Which may open a wide range of new opportunities for interaction design in immersive virtual environments.

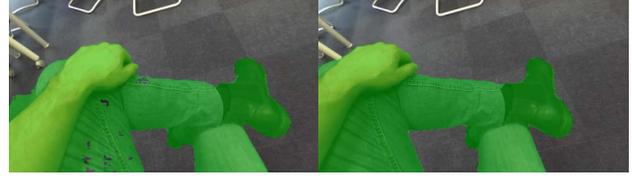

Fig. 4. Segmentation disparity between left and right images.

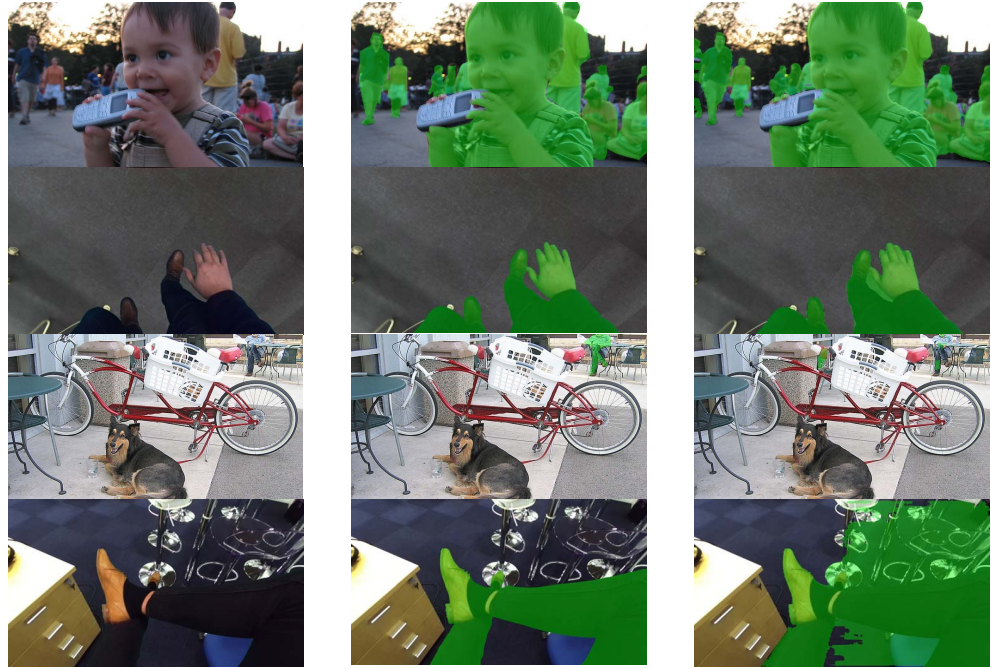

__Image #1__
IoU = 0.94
PA = 0.97
Precision = 0.95
Recall = 0.98

__Image #2__
IoU = 0.95
PA = 0.99
Precision = 0.95
Recall = 0.99

__Image #3__
IoU = 0.17
PA = 0.99
Precision = 0.48
Recall = 0.2

__Image #4__
IoU = 0.48
PA = 0.7
Precision = 0.49
Recall = 0.99

Fig. 5. Qualitative results of SSTU on COCO-body (odd rows) and EGO-body (even rows) validation sets. First column: input RGB images; second column: ground truth mask; third column: binarized SSTU outputs.

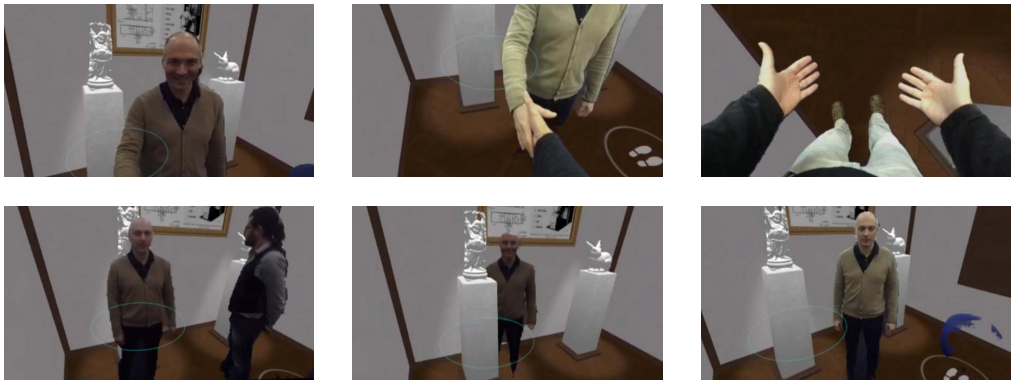

Fig. 6. Visualization of the segmented video streams composited into the rendering of a 3D environment.

## REFERENCES

[1] P. Milgram, H. Takemura, A. Utsumi, and F. Kishino, "Augmented reality: A class of displays on the reality-virtuality continuum," *Telemanipulator and Telepresence Technologies*, vol. 2351, Jan. 1994.

[2] G. Bruder, F. Steinicke, K. Rothaus, and K. Hinrichs, "Enhancing Presence in Head-Mounted Display Environments by Visual Body Feedback Using Head-Mounted Cameras," in *2009 International Conference on CyberWorlds*, 2009, pp. 43–50.

[3] H. Regenbrecht, K. Meng, A. Reepen, S. Beck, and T. Langlotz, "Mixed Voxel Reality: Presence and Embodiment in Low Fidelity, Visually Coherent, Mixed Reality Environments," in *2017 IEEE International Symposium on Mixed and Augmented Reality (ISMAR)*, 2017, pp. 90–99.

[4] C. S. Oh, J. N. Bailenson, and G. F. Welch, "A Systematic Review of Social Presence: Definition, Antecedents, and Implications," *Front. Robot. AI*, vol. 5, 2018.

[5] L. Phillips, B. Ries, M. Kaeding, and V. Interrante, "Avatar self-embodiment enhances distance perception accuracy in non-photorealistic immersive virtual environments," in *2010 IEEE Virtual Reality Conference (VR)*, 2010, pp. 115–1148.

[6] A. Pusch, O. Martin, and S. Coquillart, "Effects of Hand Feedback Fidelity on Near Space Pointing Performance and User Acceptance," *ISVRI 2011 - IEEE International Symposium on Virtual Reality Innovations 2011, Proceedings*, Mar. 2011.

[7] H. J. Snijders, N. P. Holmes, and C. Spence, "Direction-dependent integration of vision and proprioception in reaching under the influence of the mirror illusion," *Neuropsychologia*, vol. 45, no. 3, pp. 496–505, Feb. 2007.

[8] P. J. Durlach, J. Fowlkes, and C. J. Metevier, "Effect of Variations in Sensory Feedback on Performance in a Virtual Reaching Task," *Presence: Teleoperators and Virtual Environments*, vol. 14, no. 4, pp. 450–462, Aug. 2005.

[9] D. Nahon, G. Subileau, and B. Capel, "'Never Blind VR' enhancing the virtual reality headset experience with augmented virtuality," in *2015 IEEE Virtual Reality (VR)*, 2015, pp. 347–348.

[10] S. Orts-Escolano *et al.*, "Holoportation: Virtual 3D Teleportation in Real-time," in *Proceedings of the 29th Annual Symposium on User Interface Software and Technology*, New York, NY, USA, 2016, pp. 741–754.

[11] M. Camplani and L. Salgado, "Efficient Spatio-Temporal Hole Filling Strategy for Kinect Depth Maps," *Proc SPIE*, vol. 8290, p. 13, Feb. 2012.

[12] S. Khattak, B. Cowan, I. Chepurna, and A. Hogue, "A real-time reconstructed 3D environment augmented with virtual objects rendered with correct occlusion," in *2014 IEEE Games Media Entertainment*, 2014, pp. 1–8.

[13] E. K. Edwards, J. P. Rolland, and K. P. Keller, "Video see-through design for merging of real and virtual environments," in *Proceedings of IEEE Virtual Reality Annual International Symposium*, 1993, pp. 223–233.

[14] "Oculus Developer Documentation - Guidelines for VR Performance Optimization." .

[15] M. McGill, D. Boland, R. Murray-Smith, and S. Brewster, "A Dose of Reality: Overcoming Usability Challenges in VR Head-Mounted Displays," in *Proceedings of the 33rd Annual ACM Conference on Human Factors in Computing Systems*, New York, NY, USA, 2015, pp. 2143–2152.

[16] T. Günther, I. S. Franke, and R. Groh, "Aughanded Virtuality — The hands in the virtual environment," in *2015 IEEE Virtual Reality (VR)*, 2015, pp. 327–328.

[17] G. Heo, D.-W. Lee, S. Shin, G. Kim, and H. Shin, "Hand Segmentation for Optical See-through HMD Based on Adaptive Skin

Color Model Using 2D/3D Images," in *International Conference on Machine Vision and Machine Learning*, 2014.

[18] J. Maurya, R. Hebbalaguppe, and P. Gupta, "Real Time Hand Segmentation on Frugal Headmounted Device for Gestural Interface," in *2018 25th IEEE International Conference on Image Processing (ICIP)*, 2018, pp. 4023–4027.

[19] J. Tompson, M. Stein, Y. Lecun, and K. Perlin, "Real-Time Continuous Pose Recovery of Human Hands Using Convolutional Networks," presented at the ACM Transactions on Graphics, 2014, vol. 33.

[20] M. Zhao and Q. Jia, "Hand Segmentation Using Randomized Decision Forest Based on Depth Images," in *2016 International Conference on Virtual Reality and Visualization (ICVRV)*, 2016, pp. 110–113.

[21] S. Sridhar, F. Mueller, A. Oulasvirta, and C. Theobalt, "Fast and robust hand tracking using detection-guided optimization," *2015 IEEE Conference on Computer Vision and Pattern Recognition (CVPR)*, pp. 3213–3221, 2015.

[22] J. Shotton *et al.*, "Real-time human pose recognition in parts from single depth images," in *CVPR 2011*, 2011, pp. 1297–1304.

[23] F. Yang and Y. Wu, "A Soft Proposal Segmentation Network (SPS-Net) for Hand Segmentation on Depth Videos," *IEEE Access*, vol. 7, pp. 29655–29661, 2019.

[24] S. Zhao, W. Yang, and Y. Wang, "A new hand segmentation method based on fully convolutional network," in *2018 Chinese Control And Decision Conference (CCDC)*, 2018, pp. 5966–5970.

[25] O. Ronneberger, P. Fischer, and T. Brox, "U-Net: Convolutional Networks for Biomedical Image Segmentation," presented at the MICCAI 2015, 2015, vol. abs/1505.04597.

[26] J. Long, E. Shelhamer, and T. Darrell, "Fully convolutional networks for semantic segmentation," in *2015 IEEE Conference on Computer Vision and Pattern Recognition (CVPR)*, 2015, pp. 3431–3440.

[27] "Mobile Real-time Video Segmentation," *Google AI Blog*. .

[28] T.-Y. Lin *et al.*, "Microsoft COCO: Common Objects in Context," *arXiv e-prints*, p. arXiv:1405.0312, mai 2014.

[29] E. Gonzalez-Sosa, P. Perez, R. Kachach, J. J. Ruiz, and A. Villegas, "Towards Self-perception in Augmented Virtuality: Hand Segmentation with Fully Convolutional Networks," in *Proceedings of the 39th Annual European Association for Computer Graphics Conference: Posters*, Goslar Germany, Germany, 2018, pp. 9–10.

[30] "What is the latency of the ZED camera?," *Help Center | Stereolabs*. [Online]. Available: http://support.stereolabs.com/hc/en-us/articles/206918319-What-is-the-latency-of-the-ZED-camera-..

[31] "Varjo XR-1," *Varjo.com*. [Online]. Available: https://varjo.com/xr-1/. [Accessed: 26-Jul-2019].

[32] J. P. Freiwald, N. Katzakis, and F. Steinicke, "Camera Time Warp: Compensating Latency in Video See-Through Head-Mounted-Displays for Reduced Cybersickness Effects," in *2018 IEEE International Symposium on Mixed and Augmented Reality Adjunct (ISMAR-Adjunct)*, 2018, pp. 49–50.

[33] R. P. K. Poudel, S. Liwicki, and R. Cipolla, "Fast-SCNN: Fast Semantic Segmentation Network," *arXiv:1902.04502 [cs]*, Feb. 2019.

[34] W. Xiang, H. Mao, and V. Athitsos, "ThunderNet: A Turbo Unified Network for Real-Time Semantic Segmentation," in *2019 IEEE Winter Conference on Applications of Computer Vision (WACV)*, Waikoloa Village, HI, USA, 2019, pp. 1789–1796.

[35] H. Li, P. Xiong, H. Fan, and J. Sun, "DFANet: Deep Feature Aggregation for Real-Time Semantic Segmentation," in *The IEEE Conference on Computer Vision and Pattern Recognition (CVPR)*, 2019.

[36] M. Rünz, M. Buffier, and L. Agapito, "MaskFusion: Real-Time Recognition, Tracking and Reconstruction of Multiple Moving Objects," *2018 IEEE International Symposium on Mixed and Augmented Reality (ISMAR)*, pp. 10–20, 2018.